%% file: main.tex
\documentclass{article}

\usepackage[T1]{fontenc}
\usepackage{microtype}
\usepackage{graphicx}
\usepackage{subcaption}
\usepackage{booktabs}
\usepackage{amsmath}
\usepackage{amsfonts}
\usepackage{amssymb}
\usepackage{float}
\usepackage{longtable}
\usepackage{multirow}
\usepackage{pdflscape}
\usepackage{xcolor}
\usepackage{pifont}
\usepackage{tikz}

\usepackage{hyperref}

\usepackage[accepted]{icml2026}

\newcommand{\edit}[1]{#1}
\newcommand{\editdel}[1]{}
\newcommand{\xmark}{\ding{55}}
\newcommand{\cmark}{\ding{51}}

\graphicspath{{figures/}}

\icmltitlerunning{Localizing RL-Induced Tool Use to a Single Crosscoder Feature}

\begin{document}
\raggedbottom

\twocolumn[
\icmltitle{Localizing RL-Induced Tool Use to a Single Crosscoder Feature}

\icmlsetsymbol{equal}{*}
\begin{icmlauthorlist}
\icmlauthor{Andrii Shportko}{equal,nu}
\icmlauthor{Shubham Bhokare}{equal,nu}
\icmlauthor{Ahmed Zeyad A Alzahrani}{nu}
\icmlauthor{Bowen Cheng}{nu}
\icmlauthor{Gustavo Mercier}{nu}
\icmlauthor{Jessica Hullman}{nu}
\end{icmlauthorlist}
\icmlaffiliation{nu}{Northwestern University}
\icmlcorrespondingauthor{Andre Shportko}{andre.s@u.northwestern.edu}
\icmlcorrespondingauthor{Shubham Bhokare}{shubhambhokare2026@u.northwestern.edu}
\vskip 0.3in
]

\printAffiliationsAndNotice{\icmlEqualContribution}

\input{sections/abstract}
\input{sections/introduction}
\input{sections/related_work}
\input{sections/methodology}
\input{sections/feature_analysis}
\input{sections/evaluation}
\input{sections/steering}
\input{sections/discussion}
\input{sections/conclusion}

\bibliographystyle{icml2026}
\newpage
\bibliography{references}

\newpage
\appendix
\onecolumn
\input{sections/appendix}

\end{document}

%% file: sections/abstract.tex
\begin{abstract}
Fine-tuning through RL reshapes the internal representations of language models to enable agentic behaviors such as tool use, yet the mechanistic basis of these changes remains poorly understood. While RL substantially improves structured tool-call generation, it is unclear which features emerge, which are preserved, and whether identified features can be leveraged for retraining-free behavioral control. In this work, we show that \textit{Dedicated Feature Crosscoders (DFC)} isolate a compact set of RL-specific features that mediate tool-calling capability\footnote{see Limitations (section \ref{sec:discussion})} in \texttt{Qwen2.5-3B}. Across a $48$-crosscoder hyperparameter sweep, encode-decode reconstruction improves the RL model's tool correctness by $+31.1 \pm {9.7}$\,pp and passively transfers tool-calling ability to the frozen base model by $+6.8 \pm 5.0$\,pp which we call a \emph{capability spillover}. Our findings show that DFC partitioning concentrates RL-introduced capability into a minimal, steerable feature set that enables runtime behavioral control of agentic LLMs.
\end{abstract}

%% file: sections/introduction.tex
\section{Introduction}

Large language models are increasingly trained to perform agentic tasks such as invoking
external tools and interacting with external
systems~\citep{ouyang2022rlhf}. RL enables these behaviors, yet
how RL fine-tuning alters a model's internal representations is not well understood.

Mechanistic interpretability has introduced sparse autoencoders
(SAEs)~\citep{bricken2023monosemanticity,cunningham2023sparse,templeton2024scaling} to decompose model
activations into interpretable features, and Crosscoders~\citep{lindsey2024crosscoders} extend this shared decomposition across two related models. Dedicated Feature
Crosscoders~\citep{jiralerspong2025model} further partition the joint dictionary into
$A$-exclusive, $B$-exclusive, and shared sub-dictionaries with gradient masking in order to enforce
exclusivity. We test whether capability differences introduced by RL concentrate in the $A$-exclusive partition as intended and whether those features can be steered for retraining-free behavioral control.

We make the following contributions:
\begin{enumerate}
  \item We train and evaluate 48 crosscoder variants across a systematic hyperparameter sweep on a ToolRL-fine-tuned \texttt{Qwen2.5-3B} pair ($\S \ref{sec:hyperparam}$). 
  \item We identify \emph{capability spillover}: passing the frozen base model's activations through a jointly-trained crosscoder passively transfers {$+6.8$}\,pp tool-correctness with zero fine-tuning ($\S \ref{sec:spillover}$).
  \item We show that the DFC exclusive partition acts as a \emph{filter}, concentrating the strongest model-specific features but not fully isolating the RL-induced capability difference (\emph{sink})\footnote{A sink is the limiting case of a filter. It is reached only when the capability is orthogonal to shared structure.}; penalizing it degrades RL-model fidelity ($\S \ref{sec:filter}$).
  \item We demonstrate that steering a \textbf{single $A$-exclusive neuron} (interchangeably: \textit{feature}) achieves $+65.0$\,pp $\Delta$ tool-correctness, generalizing across the supermajority of probed layers ($\S \ref{sec:saturation}$).
  \item We provide geometric and autointerp evidence that $A$-exclusive features occupy a separable "Tool Interaction" region which is distinct from shared and B-exclusive features under the DFC, but not under a matched CrossCoder ($\S \ref{sec:feature_analysis}$).
\end{enumerate}

DFC-based model diffing is a promising tool for identifying and modulating the representations introduced by RL post-training. Our implications for mechanistic interpretability suggest runtime behavioral control of agentic LLMs.

%% file: sections/related_work.tex
\section{Related Work}

\begin{figure*}[t]
    \centering
    \begin{subfigure}[b]{0.32\linewidth}
        \centering
        \includegraphics[width=\linewidth]{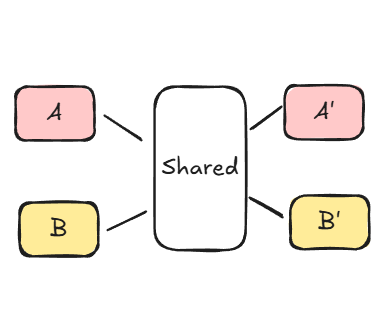}
        \caption{Crosscoders}
        \label{fig:panel-a}
    \end{subfigure}
    \hfill
    \begin{subfigure}[b]{0.32\linewidth}
        \centering
        \includegraphics[width=\linewidth]{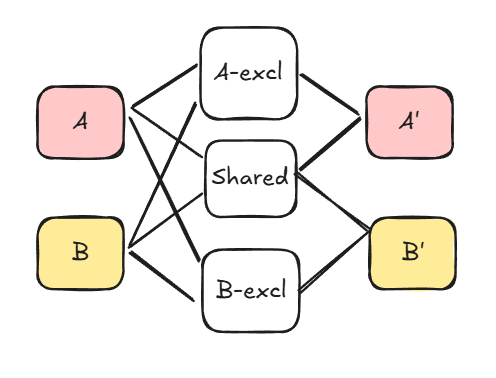}
        \caption{Tied Crosscoders}
        \label{fig:panel-b}
    \end{subfigure}
    \hfill
    \begin{subfigure}[b]{0.32\linewidth}
        \centering
        \includegraphics[width=\linewidth]{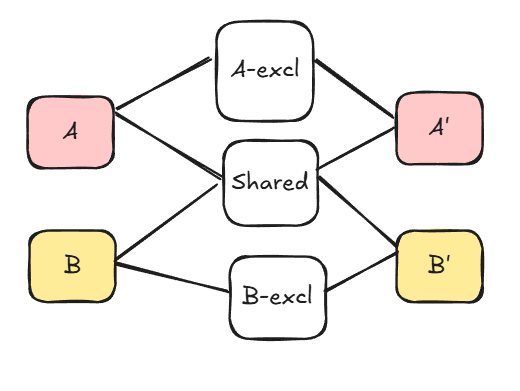}
        \caption{Dedicated Feature Crosscoders}
        \label{fig:panel-c}
    \end{subfigure}
    \caption{Three architectures for joint sparse decomposition of paired model activations}
    \label{fig:three-panel}
\end{figure*}

\paragraph{Sparse autoencoders and mechanistic interpretability.}
SAEs are popular as a tool for decomposing LLM activations into interpretable features~\citep{bricken2023monosemanticity,cunningham2023sparse,templeton2024scaling,elhage2022superposition}.
These methods are built on the hypothesis that model representations are superpositions of many more features than there are neurons, and that sparse dictionary learning can recover them.

\paragraph{Crosscoders and model diffing.}
Crosscoders~\citep{lindsey2024crosscoders} (fig. \ref{fig:panel-a}) extend SAEs to pairs of models by jointly
encoding their activations into one shared sparse dictionary. Dedicated Feature
Crosscoders~\citep{jiralerspong2025model} (fig. \ref{fig:panel-c}), like the Tied Crosscoders ~\citep{aranguri2025tied} (fig. \ref{fig:panel-b}), further partition this dictionary with gradient
masking. It forces each model to read from and write to only its designated partition.
Our work is the first systematic behavioral evaluation of DFC across a hyperparameter
sweep, and the first to demonstrate single-neuron steering saturation.

\paragraph{RL fine-tuning for tool use.}
ToolRL~\citep{acikgoz2025toolrl} demonstrates that RL from execution feedback
substantially improves structured tool-call generation in \texttt{Qwen2.5-3B}. We use the
ToolRL-fine-tuned model as Model~$A$ throughout our experiments.

\paragraph{Activation steering.}
Our steering approach operates in the sparse
dictionary of the crosscoder which allows for feature-level intervention \cite{templeton2024scaling}. We identify the minimal
steerable subset of neurons, and a single $A$-exclusive neuron is enough to maximize the effect.

\paragraph{Automated interpretability.}
Automated interpretability methods~\citep{bills2023language} use
LLM judges to label and score SAE features. We apply the protocol
to validate $A$-exclusive feature interpretations.

%% file: sections/methodology.tex
\section{Methodology}
\label{sec:methodology}

\subsection{Models and Training Data}

We study two models sharing the \texttt{Qwen2.5-3B} architecture ($d=2048$, $36$ layers):
\textbf{Model B (Base):} \texttt{Qwen/Qwen2.5-3B}~\citep{qwen2025} and
\textbf{Model A (RL):} \texttt{chengq9/ToolRL-Qwen2.5-3B}~\citep{acikgoz2025toolrl},
RL-fine-tuned for structured \texttt{<tool\_call>} invocation.

Training uses $40{,}000$ \texttt{FineWeb} \citep{penedo2024fineweb} samples (general-domain) and
$40{,}000$ \texttt{ToolRL} instruction-output pairs.

\subsection{DFC Architecture and Training Objective}

The DFC dictionary of size $D$ is partitioned into A-exclusive $[0, a_\text{end})$,
B-exclusive $[a_\text{end}, b_\text{end})$, and shared $[b_\text{end}, D)$ features,
with gradient masking enforcing exclusivity. The training objective is presented in the Equation [\ref{eq:loss}]:
\begin{equation}
\begin{split}
\mathcal{L} = {}& \mathrm{MSE}(h, \hat{h}) + \lambda_{\mathrm{sh}} \cdot \overline{\|f_{\mathrm{sh}}\|_1} \\
& + \frac{\lambda_{\mathrm{excl}}}{2}\left( \overline{\|f_{A\cup\mathrm{sh}}\|_1} + \overline{\|f_{B\cup\mathrm{sh}}\|_1} \right)
\end{split}
  \label{eq:loss}
\end{equation}
where $h=(h_A,h_B)$ is the concatenated residual stream post-MLP and $\overline{|\cdot|_1}$ denotes the mean of $|f_i|$. Top-$k$
sparsity is enforced in the encoder. 

\subsection{Hyperparameter Sweep}
\label{sec:hyperparam}
The sweep varies five axes that determine the decomposition's capacity, sparsity, and partition structure (Table \ref{tab:sweep_config}). \textbf{Architecture} is the top-level choice between a standard Crosscoder (one unpartitioned dictionary jointly encoding both models) and a DFC (the same dictionary explicitly split into A-exclusive, B-exclusive, and shared sub-dictionaries via gradient masking). \textbf{Dictionary size} $D$ is the total number of features the dictionary can represent. \textbf{Top-$k$} is the sparsity budget: at each forward pass, only the $k$ highest-activating features are retained. \textbf{Exclusive share} $p$ (DFC only) is the fraction of $D$ allocated to each exclusive partition. $\boldsymbol{\lambda_{\text{excl}}}$ {is the magnitude penalty applied specifically to the exclusive partitions (in the CrossCoder rows of Table~\ref{tab:full_sweep_eval} we reuse the same coefficient on the equivalent unmasked features --- see code release):} setting it to $0$ (``free'') lets exclusive features fire without extra cost, while $10^{-3}$ (``penalised'') pushes the model to route signal through shared features unless exclusivity is needed which allows to test whether the exclusive partition behaves as a sink or a filter. 

\begin{table}[t]
\centering
\small
\begin{tabular}{@{}p{0.32\columnwidth}p{0.60\columnwidth}@{}}
\toprule
\textbf{Axis} & \textbf{Values} \\
\midrule
Architecture    & CrossCoder (no exclusive partition), DFC \\
Dictionary $D$  & 8{,}192 \quad 16{,}384 \\
Top-$k$         & 45 \quad 90 \quad 160 \\
Exclusive share $p$ (DFC) & 3\% \quad 5\% \quad 10\% \\
$\lambda_\text{excl}$ {(DFC; reused for CC variants)} & 0 (free) \quad $10^{-3}$ (penalised) \\
\bottomrule
\end{tabular}
\caption{$48$ crosscoder variants ($36$ DFC and $12$ crosscoders) trained for $9{,}000$ steps, batch $1{,}024$,
         Adam $lr=10^{-4}$, $\lambda_\text{sh}=10^{-3}$.}
\label{tab:sweep_config}
\end{table}

\subsection{Behavioral Scoring Rubric}

Each generated response is scored on three metrics~{\citep{acikgoz2025toolrl}}:
\begin{itemize}
    \item \textbf{format\_accuracy}: \texttt{<tool\_call>} present and JSON \texttt{"name":"..."} field present;
    \item  \textbf{tool\_correctness}: called name fuzzy-matches a numbered tool in
the prompt; 
    \item \textbf{overall\_score} $\in\{-1,0,+1,+2\}$: $+2$ = both correct,
$0$ = format only, $+1$ = tool only, $-1$ = neither. Evaluation: 100 held-out ToolRL
prompts per sweep variant and 40 prompts per steering cell (seed 42), greedy decoding,
\texttt{max\_new\_tokens=200}, \texttt{max\_length=2048}.
\end{itemize}

\subsection{Neuron Identification and Targeted Steering}

Features are ranked by Cohen's $d$ on tool-use vs.\ general-text activations [Eq. \ref{eq:cohend}]:
\begin{equation}
  d_i = \frac{\mu^\text{tool}_i - \mu^\text{gen}_i}
             {\sqrt{(s^{\text{tool}\,2}_i + s^{\text{gen}\,2}_i)/2}}
  \label{eq:cohend}
\end{equation}
and filtered by firing rate ($\rho^\text{tool}_i\geq0.3$, $\rho^\text{gen}_i\leq0.1$).
Targeted steering applies an additive correction to Model A's residual stream [Eq. \ref{eq:steer}]:
\begin{equation}
  h'_A = h_A + \sum_{i \in S} (\alpha-1)\cdot f_i \cdot W_\text{dec}[i,A,:]
  \label{eq:steer}
\end{equation}
for subset $S$ of A-exclusive features ranked by $d_i$, steering coefficient $\alpha$,
and decoder column $W_\text{dec}[i,A,:]\in\mathbb{R}^d$.
(Section~\ref{sec:evaluation}).

%% file: sections/feature_analysis.tex
\section{Feature Space Analysis}
\label{sec:feature_analysis}

\subsection{DFC vs CrossCoder Decoder Geometry}
\label{sec:umap}
\begin{figure*}[t]
    \centering
    \includegraphics[width=1\linewidth]{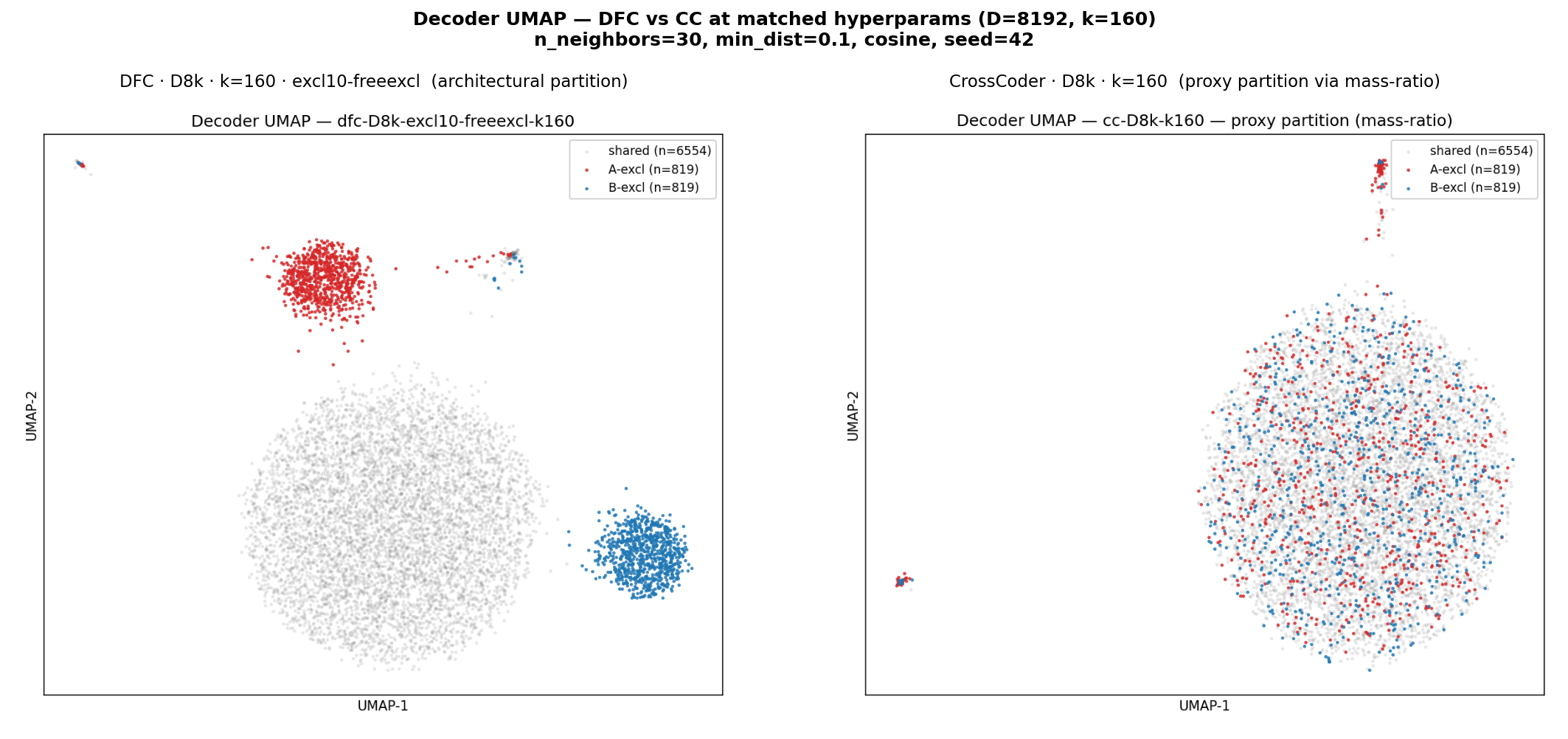}
    \caption{Decoder UMAP at matched hyperparameters ($D = 8{,}192$, $k = 160$; $n_{\text{neighbors}} = 30$, $\min\_\text{dist} = 0.1$, cosine metric, seed $42$). \textbf{Left:} the DFC (\texttt{dfc-D8k-excl10-freeexcl-k160}) yields three spatially distinct regions: A-exclusive (red), B-exclusive (blue), and shared (grey). \textbf{Right:} the CrossCoder (\texttt{cc-D8k-k160}) with a mass-ratio proxy partition of identical sizes ($819 / 819 / 6{,}554$) shows A-biased and B-biased features mixing uniformly with shared features throughout the dense core.}
    \label{fig:umap_partition}
\end{figure*}

To test whether the partitioned geometry observed for the DFC (Figure \ref{fig:umap_partition}) is a consequence of the gradient-mask architecture or merely of label imbalance, we construct a matched-size proxy partition for the unconstrained CrossCoder. For each dictionary feature $i$, the decoder has two columns: $W_{\text{dec}}[i, 0, :]$, the direction the feature decodes into Model A's residual stream, and $W_{\text{dec}}[i, 1, :]$, the direction it decodes into Model B's. We define the \emph{mass-ratio}
\begin{equation}
r_i = \log\!\frac{\lVert W_{\text{dec}}[i, 0, :]\rVert_2}{\lVert W_{\text{dec}}[i, 1, :]\rVert_2}
\end{equation}
so that $r_i \gg 0$ indicates a feature writing almost entirely into Model A (effectively A-exclusive), $r_i \ll 0$ indicates Model B (effectively B-exclusive), and $r_i \approx 0$ indicates balanced contribution (shared). For the DFC, the partition mask forces this ratio: the first $819$ features have $W_{\text{dec}}[i, 1, :] = 0$ (so $r_i \to +\infty$), the next $819$ have $W_{\text{dec}}[i, 0, :] = 0$ ($r_i \to -\infty$), and the remaining $6{,}554$ have unconstrained A and B decoders. For the CrossCoder no such constraint exists, so we sort all $8{,}192$ features by $r_i$ and label the top $819$ as the A-exclusive proxy, the bottom $819$ as the B-exclusive proxy, and the middle $6{,}554$ as the shared proxy. This produces partition slices of identical size to the DFC's, so any difference in UMAP separation is attributable to the training objective (partition mask vs no mask) rather than to label imbalance.

With matched sizes, the CrossCoder's A-biased features still mix uniformly with B-biased and shared features. The A-exclusive proxy in the CC has \emph{negative} silhouette ($-0.168$) --- on average, mass-ratio-A-biased features sit closer to non-A-biased centroids than to their own --- and $k$-NN purity drops from the DFC's $0.984$ to $0.158$, {i.e.\ the precision of recovering A-exclusive proxy labels among an A-biased feature's nearest neighbours falls to roughly the prior $|A|/D \approx 10\%$, the chance level for a $10\%$ minority class under random label assignment}. HDBSCAN clusters in the CC UMAP recover the mass-ratio labels at ARI $= 0.08$ versus $0.93$ for the DFC: the structure simply isn't there to be recovered. This falsifies the alternative explanation that the DFC's clean separation is a UMAP artefact: with matched dictionary size ($D = 8{,}192$) and matched sparsity ($k = 160$), the same UMAP pipeline produces no separation when the partition mask is removed. The geometric structure in the DFC plot is caused by the architectural partitioning and mass-asymmetry alone is insufficient to produce it. The gradient mask reorganises feature directions, not just decoder magnitudes.

%% file: sections/evaluation.tex
\section{Reconstruction Fidelity and Capability Spillover}
\label{sec:evaluation}

\subsection{Sweep-Level Reconstruction}

\begin{table*}[!t]
\centering
\begin{tabular*}{\textwidth}{@{\extracolsep{\fill}}lrr}
\toprule
\textbf{Metric} & \textbf{Model A (ToolRL)} & \textbf{Model B (Base)} \\
\midrule
Pre-recon \texttt{tool\_correctness}  & 19\%   & 0\%  \\
Post-recon \texttt{tool\_correctness} & 50.1\% & {$6.8$\%} \\
$\Delta$ tool-correctness             & $+31.1\pm{9.7}$\,pp & {$+6.8\pm5.0$}\,pp \\
$\Delta$ overall\_score               & {$+0.93$} & {$+0.14$} \\
$\Delta$ format\_accuracy             & substantial & $+0$\,pp identically \\
\bottomrule
\end{tabular*}
\caption{Sweep mean $\pm$ std, $n=48$ crosscoders, $100$ hold-out prompts each.}
\label{tab:sweep_results}
\end{table*}

Across the $48$-model sweep, reconstruction substantially improves the RL model's
behavioral performance relative to its pre-reconstruction baseline (Table \ref{tab:sweep_results}). {Every one of the $48$ trained variants improves Model~$A$ tool correctness ($48 / 48$); under a one-sided exact-binomial sign test against the chance rate of $0.5$, this corresponds to $p \approx 3.6 \times 10^{-15}$.}

At the same time, behavioral preservation (and even performance boost) does not track raw reconstruction loss in
any simple way. Across the sweep, training MSE correlates only weakly with the
gain in Model A performance ($r=+0.08$, 95\% CI $[-0.21,+0.36]$). In other words, lower MSE does not reliably predict better behavioral reconstruction. This suggests
that the information required for successful tool calling occupies a relatively sparse
and behaviorally privileged subspace: a model can incur substantial activation-level
distortion while still preserving the task-relevant structure needed to emit correct
tool calls. One plausible interpretation is that the top-$k$ bottleneck acts as an
implicit regulariser, discarding noisy or behaviorally irrelevant components while
retaining the sparse features most important for tool-use performance.

\subsection{Capability Spillover}
\label{sec:spillover}

Model B's tool-correctness rises from 0\% to {$6.8\%$} post-reconstruction {(paired-$t$ across the $48$ runs against the always-zero baseline: $t \approx 9.4$, $p \approx 1\times10^{-12}$)} without any fine-tuning. We call this \emph{capability spillover}:
tool-calling intent is delocalized into the shared decoder weights during joint training
and reconstruction passively routes it into Model B's activation stream.

\textbf{Format accuracy never spills over} ($\Delta_B$ format $=0$\,pp across all 48 runs).
Model B correctly identifies tools in prose but never produces the exact
\texttt{<tool\_call>..."name":...} surface form. It
suggests that semantic tool-selection intent is at least partially represented in shared
directions, but the exact surface-form machinery required to produce the
\texttt{<tool\_call>} template remains isolated in narrow RL-specific features. This
interpretation is consistent with the later autointerp evidence, which shows that the
most discriminative A-exclusive features are structural-template detectors rather than
broad abstract tool-use concepts.

\subsection{Architecture Comparison}

\begin{table}[t]
\centering
\small
\begin{tabular}{lrrrr}
\toprule
\textbf{Arch.} & $n$ & $\Delta A$ (pp) & $\Delta B$ (pp) & Train MSE \\
\midrule
CrossCoder & 12 & $+32.9$ & $+8.9$ & $0.031\pm0.010$ \\
DFC        & 36 & {$+30.5$} & {$+6.1$} & $0.031\pm0.007$ \\
\bottomrule
\end{tabular}
\caption{Welch's $t$-test on $\Delta_B$: $t=1.63$, $p=0.12$ (directional, not
         significant). Train MSE indistinguishable.}
\label{tab:arch}
\end{table}

Comparing the two architectures at sweep level (Table \ref{tab:arch}), CrossCoders and DFCs achieve
essentially indistinguishable reconstruction error but somewhat different behavioral
transfer profiles. CrossCoders ($12$ runs) achieve a mean Model A tool-correctness
gain of $+32.9$\,pp and a mean Model B gain of $+8.9$\,pp, while DFCs (36 runs)
achieve {$+30.5$\,pp on Model A and $+6.1$\,pp on Model B}. Mean training MSE is
the same to three decimal places for both families. Raw
reconstruction loss alone does not explain the behavioral differences.

The directional reduction in Model B spillover under DFC is suggestive but not
statistically decisive in this sweep. A Welch's $t$-test on $\Delta_B$ yields
$t=1.63$, $p=0.12$, so we do not claim a significant architecture-level difference
here. The more cautious interpretation is that DFC \emph{may} reduce spillover
relative to an unpartitioned CrossCoder, but the evidence is currently underpowered
and should be treated as directional rather than conclusive.

\subsection{The DFC Partition as a Filter, Not a Sink}
\label{sec:filter}
{$\lambda_{\text{excl}} = 10^{-3}$ reduces Model~$A$ fidelity relative to
$\lambda_{\text{excl}} = 0$ at both the $5\%$ and $10\%$ exclusive shares
($34.8 \to 25.8$\,pp $\Delta A$ at $5\%$ and $35.2 \to 32.0$\,pp at $10\%$,
each averaged over the six matched $(D,k)$ pairs in
Appendix~\ref{app:full_eval}, Table~\ref{tab:full_sweep_eval}).
Sparsifying the exclusive partition therefore pushes tool-specific signal back into the shared partition rather than removing it.} Rather than isolating all RL-specific signal into a protected compartment, the exclusive partition appears to act more like a \emph{filter} that concentrates the most
model-specific residue.

If the exclusive slice were a true sink for the capability
difference, sparsifying it should mostly remove redundant or nonessential signal while leaving shared structure intact. Instead, penalizing the exclusive slice appears to push some tool-relevant information back into the shared partition, where it is both less behaviorally efficient for Model $A$ and more available for spillover into Model $B$.
The DFC partition therefore helps localize model-specific features, but it does not perfectly disentangle capability-specific content from the shared representational substrate.

%% file: sections/steering.tex
\section{Targeted-Neuron Steering}
\label{sec:steering}



\subsection{DFC Saturates at a Single Feature}
\label{sec:saturation}

\vspace{-4pt}
\enlargethispage{\baselineskip}
\begin{figure}[H]
\centering
\includegraphics[width=\linewidth]{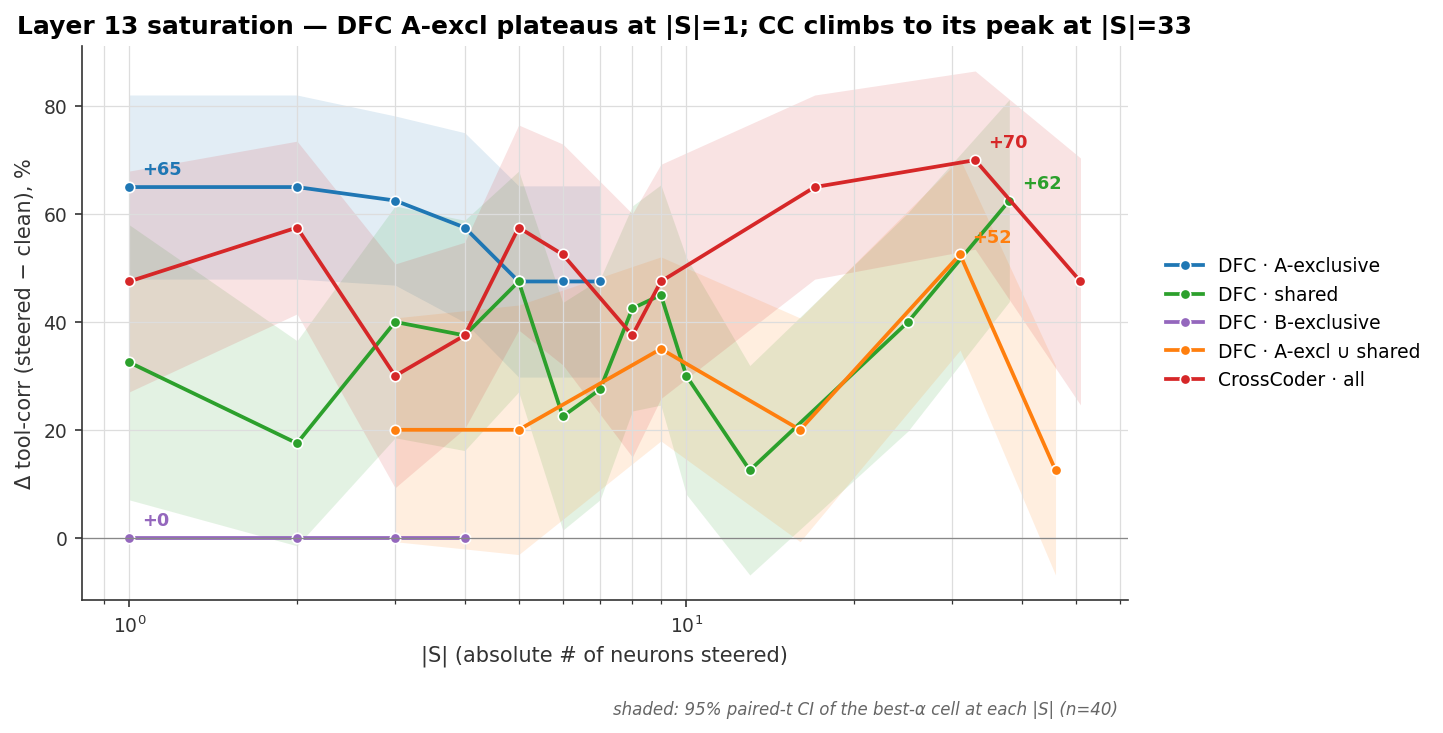}
\caption{Saturation curve: $\Delta$ tool-correctness vs $|S|$ (log-$x$) for all
         conditions at layer 13. DFC A-excl plateaus from $|S|=1$; CrossCoder needs
         $|S|=33$ to reach its peak. B-excl steering has zero effect throughout. Devices used: \texttt{dfc-D8k-excl10-k45} and \texttt{cc-D8k-k45}}
\label{fig:saturation}
\end{figure}
\vspace{-4pt}

Steering a \textbf{single A-exclusive neuron} ($|S|=1$, $\alpha=32$) achieves
$\mathbf{+65.0}$\,pp $\Delta$ tool-correctness (95\% CI $[+47.9,+82.1]$), matching
the performance of steering all 7 available A-exclusive neurons. CrossCoder requires
$|S|=33$ to reach its unbudgeted peak of $+70.0$\,pp ($[+53.5,+86.5]$) (Fig. \ref{fig:saturation}).
DFC A-exclusive plateaus immediately; CC climbs slowly across $33\times$ the budget. The full per-$|S|$ breakdown with CIs is in Table~\ref{tab:saturation_full} (Appendix~\ref{app:saturation_table}).

\subsection{Ablation Studies}

Table~\ref{tab:steering_best} isolates four mechanisms. B-exclusive steering has zero effect at every cell. A-exclusive beats shared under a small budget, despite shared holding more tool-relevant features by Cohen's $d$. Signal in A-exclusive is concentrated; signal in shared is diffuse. Steering A-exclusive and shared together underperforms A-exclusive alone. The likely cause is destructive interference between non-orthogonal decoder directions amplified at matched $\alpha$. The unpartitioned CrossCoder eventually matches DFC A-exclusive's peak, but only by recruiting an order of magnitude more features.
\vspace{-4pt}
\begin{table}[H]
\centering
\small
\resizebox{\columnwidth}{!}{%
\begin{tabular}{lrrrl}
\toprule
\textbf{Condition} & \textbf{Best $\Delta$ (\%)} & \textbf{95\% CI} & $|S|$ & $\alpha$ \\
\midrule
\multicolumn{5}{l}{\textit{Within $|S|\leq10$}} \\
\textbf{DFC $\cdot$ A-exclusive}  & \textbf{+65.0} & [+47.9,+82.1] & \textbf{1} & 32 \\
DFC $\cdot$ shared                 & +47.5          & [+27.0,+68.0] & 5           & 16 \\
DFC $\cdot$ B-exclusive            & +0.0           & [+0.0,+0.0]   & 1           & 1  \\
DFC $\cdot$ A-excl $\cup$ shared  & +35.0          & [+17.9,+52.1] & 9           & 16 \\
CrossCoder $\cdot$ all             & +57.5          & [+41.5,+73.5] & 2           & 32 \\
\midrule
\multicolumn{5}{l}{\textit{Unbudgeted (any $|S|$)}} \\
DFC $\cdot$ A-exclusive            & +65.0          & [+47.9,+82.1] & 1           & 32 \\
DFC $\cdot$ shared                 & +62.5          & [+43.8,+81.2] & 38          & 6  \\
\textbf{CrossCoder $\cdot$ all}   & \textbf{+70.0} & [+53.5,+86.5] & 33         & 6  \\
\bottomrule
\end{tabular}}
\caption{Best targeted-steering cell per condition, layer 13, $n=40$ prompts per cell.
         DFC A-excl achieves its maximum at $|S|=1$; CC requires $|S|=33$.}
\label{tab:steering_best}
\end{table}

\subsection{DFC vs CrossCoder: Budget-Dependent Advantage}
\label{sec:budget}


DFC A-exclusive significantly outperforms shared, B-exclusive, and the A-excl $\cup$ shared combination, with effect sizes ranging from moderate ($d_z = 0.40$) to very large ($d_z = 1.03$) (Table~\ref{tab:paired}). Against the unconstrained CrossCoder, the difference is null ($d_z = 0.00$, $p = 0.51$). {Within the power of this study, the two architectures' unbudgeted ceilings are indistinguishable.} This is the central tradeoff. {The DFC's value is that it reaches \emph{its} ceiling with one neuron, whereas the CrossCoder needs thirty-three to reach \emph{its}.} For interpretability and targeted intervention, single-neuron control is qualitatively different from distributed control; for raw behavioral effect at any cost, the architectures are interchangeable.

\vspace{-4pt}
\begin{table}[H]
\centering
\small
\resizebox{\columnwidth}{!}{%
\begin{tabular}{lrrr}
\toprule
\textbf{Comparison} & \textbf{Mean $\Delta$ (pp)} & Cohen's $d_z$ & $p$ (one-sided) \\
\midrule
A-excl $-$ shared & +10.5   & 0.40 & \textbf{0.019} \\
A-excl $-$ B-excl & +24.3   & 1.03 & \textbf{$2.2\times10^{-6}$} \\
A-excl $-$ combo  & +15.7   & 0.56 & \textbf{0.002} \\
A-excl $-$ CC     & $-$0.08 & 0.00 & 0.51 (n.s.) \\
\bottomrule
\end{tabular}}
\caption{Cell-level paired tests vs DFC A-excl ($n=30$ shared cells, one-sided
         H1: A-excl $>$ baseline). A-excl beats shared, B-excl, and combo;
         ties CC unbudgeted}
\label{tab:paired}
\end{table}

\subsection{Cross-Layer Generalisation}
\label{sec:crosslayer}

\vspace{-4pt}
\enlargethispage{\baselineskip}
\begin{figure}[H]
\centering
\includegraphics[width=0.82\linewidth]{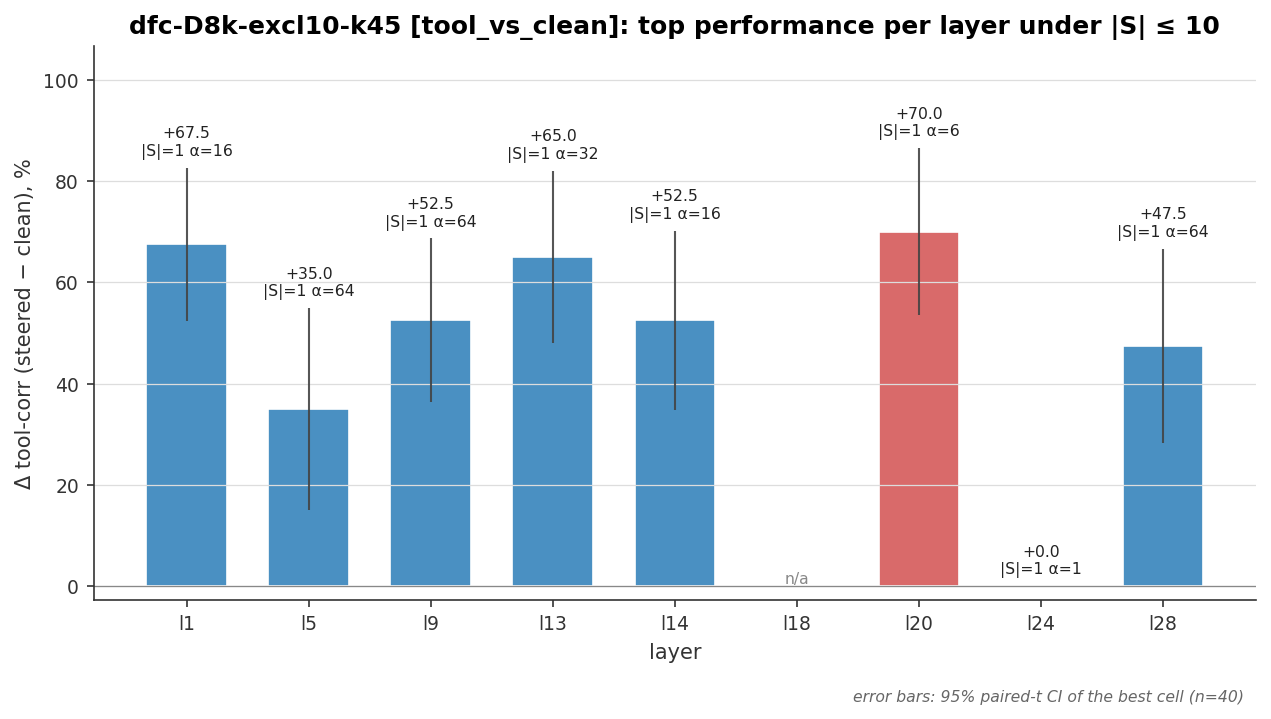}
\caption{\% tool-correctness improvement per layer at $|S|\leq10$
         (DFC A-excl, \texttt{dfc-D8k-excl10-k45}).
         Layers 18 and 24 are dead (no tool-signed A-excl features).}
\label{fig:cross_layer}
\end{figure}

Across-layers (fig. \ref{fig:cross_layer}), the mean best-cell is $\Delta=+43.3$\,pp
(95\% CI $[+22.7,+64.0]$), Cohen's $d_z=1.61$,
and paired $t$ $p=0.0006$, Wilcoxon $p=0.0078$.
Single-neuron saturation is not a layer-$13$ artefact.

\subsection{B-Exclusive Steering Has No Effect}
\label{sec:bexcl}

B-exclusive steering produces $\Delta=+0.0$\,pp at every $(|S|,\alpha)$ cell
(CI $[+0.0,+0.0]$). {We include this comparison as an end-to-end sanity check on the steering harness rather than a behavioural baseline: by construction, B-exclusive decoder columns have their Model-A output zeroed by the gradient mask, so any non-zero effect here would indicate a bug in our patching pipeline (wrong layer, wrong projection, leaked gradient). The flat $+0.0$\,pp result therefore confirms that the harness is working as intended, and the magnitude of the A-excl effect ($d_z=1.03$, $p=2.2\times10^{-6}$, Table~\ref{tab:paired}) is being read out faithfully.}

\subsection{Combining A-excl and Shared Causes Interference}
\label{sec:interference}

Steering A-excl $\cup$ shared simultaneously is worse than either alone: the best combo
$+35.0$\,pp at $|S|=9$ vs A-excl $+65.0$\,pp at $|S|=1$ {(prompt-paired one-sided $t$ across the $n=40$ ToolRL prompts shared by both cells, $p=0.002$, Cohen's $d_z=0.56$; the $30$ shared $(|S|,\alpha)$ cells in Table~\ref{tab:paired} aggregate these prompt-level paired tests across cells)}. The most likely mechanism is destructive interference between non-orthogonal
A-excl and shared decoder directions boosted at equal $\alpha$.

\subsection{Autointerp of the most important neuron}
\label{sec:steer_autointerp}


$A$-exclusive features (such as feat. $\#136$ from figure \ref{fig:toolrl_before_after}) are \emph{structural-template detectors}. They fire on narrow formatting markers (\texttt{<tool\_call>}, \texttt{<response>}, parameter boilerplate). The max-activating fragment receives the explanation (pseudocode from Appendix \ref{app:autointerp}): \emph{``the structure of a dialogue system, specifically the interaction between a tool call and a subsequent response.''} DFC partitioning concentrates this detector in A-exclusive parameter space. Cohen's $d$ values larger than $\sim 5$ here reflect a near-zero variance under the general-text condition (the feature simply does not fire there) and should be read as ``cleanly separated'' rather than as comparable effect sizes; the AUROC numbers, as used in Appendix~\ref{app:autointerp_table} for another device, are the more meaningful ranking signal.

\subsection{Hypotheses Summary}
\label{sec:hypotheses}
Please see Table \ref{tab:hypotheses}.

\vspace{-4pt}
\begin{table*}[t]
\centering
\small
\begin{tabular}{cllll}
\toprule
\textbf{H} & \textbf{Claim} & \textbf{Status} & \textbf{Key evidence} & \textbf{Section} \\
\midrule
H1 & Steering helps overall            & \cmark\ confirmed   & {Cross-layer; paired-$t$ $p=0.0006$, Wilcoxon $p=0.0078$} & \S\ref{sec:crosslayer} \\
H2 & A-excl $\gg$ B-excl              & \cmark\ confirmed   & $d_z=1.03$, $p=2.2\times10^{-6}$       & \S\ref{sec:bexcl} \\
H3 & A-excl $>$ shared at small $|S|$ & \cmark\ confirmed &
{Pooled $d_z{=}0.40$, $p{=}0.019$} & \S\ref{sec:budget} \\
H4 & DFC $\not>$ CC unbudgeted         & \cmark\ confirmed   & $p=0.51$                                & \S\ref{sec:budget} \\
H5 & DFC $>$ CC at small $|S|$         & \cmark\ confirmed   & $|S|=1$ gap $+17.5$\,pp                & \S\ref{sec:budget} \\
H6 & DFC saturates faster              & \cmark\ confirmed   & A-excl plateaus $|S|=1$; CC needs 33   & \S\ref{sec:saturation} \\
H7 & Combo $>$ either alone            & \xmark\ rejected    & Best combo $+35$ vs A-excl $+65$        & \S\ref{sec:interference} \\
H8 & Layer-13 generalises              & \cmark\ confirmed   & 7/9 layers $\geq+35$\,pp at $|S|=1$   & \S\ref{sec:crosslayer} \\
\bottomrule
\end{tabular}
\caption{Hypotheses and status from the cross-layer and ablation sweeps.}
\label{tab:hypotheses}
\end{table*}

%% file: sections/discussion.tex
\section{Discussion}
\label{sec:discussion}

\paragraph{Joint training delocalises capability.}
Our results partially refute the DFC's design intent: that the exclusive partition cleanly contains the
model-specific capability. Tool-calling capability survives in
the shared decoder weights well enough to result in {$+6.8$}\,pp spillover to Model B and to
make $\lambda_\text{excl}$ penalisation hurt Model A fidelity. This is empirical
evidence that superposition crosses model boundaries in joint sparse decompositions when
underlying capabilities are not orthogonal in residual-stream space.

\paragraph{A minimal feature set carries most of the signal.}
UMAP+HDBSCAN finds tens of clustered features out of 819 A-exclusive; the discrimination ranking concentrates Cohen's $d$ in single-digit feature indices; and a single feature suffices for maximum steering effect across 7 of 9 probed layers. This suggests that sparse-feature-level
interpretability of agentic capabilities works.

\paragraph{Capability transfer is a side channel.}
The fact that a frozen base model gains {$+6.8$}\,pp tool-correctness purely from the decomposition pipeline, with no fine-tuning, suggests that releasing crosscoders trained between model pairs with different capability levels may inadvertently provide a substrate that reintroduces the capability at inference time. 

\paragraph{Steering as low-intervention behavioral control.}
We find that $+65$\,pp tool-correctness from a post-top-$k$ delta on a single feature requires zero
retraining. The same machinery can in principle suppress unwanted behaviors by clamping the corresponding features to zero, positioning DFC-based steering alongside RLHF/DPO/SFT for safety post-training with the advantages of being inference-time, gradient-free, and
inspectable.

\paragraph{Limitations.}
We use ``capability'' as shorthand for reliable, structured
tool-call generation. We do not claim RL installs a tool-use ability the
base model wholly lacks. Our metrics measure the \emph{propensity} to emit
correct tool calls under a fixed evaluation prompt, and we did not run a
prompted baseline tuned to maximally elicit base-model tool use; our
interventions should be read as modulating this propensity.
We study a single model pair (Qwen2.5-3B) and a single task
(ToolRL). The architecture-effect $t$-test is underpowered {(12 CC vs 36 DFC runs)}. The steering results come from $n=40$ prompts per cell; the best-cell numbers have not been
replicated on a fresh sample. The autointerp pass covered only A-exclusive features (Section~\ref{sec:steer_autointerp}); the shared partition is the natural next target given that it carries most of the spilled tool capability.

\paragraph{Future Work.} Since RL shapes agents that take consequential actions, the behaviors RL installs should become objects we can monitor and steer. Model diffing is a step toward that because if RL-induced behaviors reliably condense into small feature sets, then steering them offers runtime control that complements RLHF, DPO, and SFT, with the advantage of being legible. Realising this vision will depend on having robust diffing methods; \emph{Delta-Crosscoders}
\citep{kassem2026delta}, which explicitly model the asymmetric nature of fine-tuning, are a promising path toward the fidelity such control would
require. A further frontier is \emph{knowledge-boundary discovery}
\citep{li2024knowledge}: the most consequential thing an agent decides is
often whether to act at all. Locating the features that govern the
boundary between committing to an action and abstaining under uncertainty would let us inspect an agent's willingness to act.

%% file: sections/conclusion.tex


\section*{Acknowledgments}
We appreciate Prof. Manling Li's guidance on the task selection and future work directions. We are thankful to Thomas Jiralerspong for clarifying our questions about the architecture and setup. We are grateful to Yuliia Tkachuk for her help with the visualizations (Fig. \ref{fig:three-panel}). This work was supported in part by the Northwestern University Department of Computer Science. We also thank Prof. Hullman for additional support.

%% file: sections/appendix.tex
\section{Autointerp Pseudocode}
\label{app:autointerp}

\begin{verbatim}
AUTOINTERP(features, texts, K=10):
  # Stage 1: Find top-K activating examples per feature
  for each shard in feature_cache:
    for each feature f:
      maintain min-heap of size K by activation value

  # Stage 2: Explain
  for each feature f where topk[f] is non-empty:
    top_texts = lookup texts for topk[f]
    explanation = LLM("what pattern do these texts share?", top_texts)

  # Stage 3: Score (detection task)
    positives = top_texts              # should match explanation
    negatives = sample K random texts  # should not match
    combined  = shuffle(positives + negatives)
    predictions = LLM("MATCH or NO_MATCH?", explanation, combined)
    score = accuracy(predictions, ground_truth)
    interpretable = score >= 0.8
\end{verbatim}

\section{Top A-Exclusive Feature Interpretations}
\label{app:autointerp_table}

From \texttt{dfc-D8k-excl10-freeexcl-k160}, $n=40$ prompts, ToolRL test split.

\begin{table}[H]
\centering
\small
\begin{tabular}{rrrrl}
\toprule
\textbf{Feat} & \textbf{Cohen's $d$} & \textbf{AUROC} & \textbf{Det.} &
\textbf{Gemma explanation} \\
\midrule
521 & 12.18 & 0.998 & 0.80 & LM instructed to use tools for user requests \\
730 & 4.80  & 0.954 & 0.50 & Retrieving info via unique identifiers \\
538 & 4.43  & 0.949 & 0.80 & Instructions/guidelines for tool calls \\
126 & 4.12  & 0.929 & 0.75 & Referencing previous conversation turns \\
636 & 2.36  & 0.826 & 0.55 & LM interacting with tools and external APIs \\
17  & 2.12  & 0.901 & 0.50 & Structured data input in dialogue systems \\
583 & 1.92  & 0.882 & 0.70 & Structure and syntax of \texttt{<tool\_call>} \\
786 & 1.66  & 0.892 & 0.70 & Requests requiring external tools \\
10  & 0.90  & 0.635 & 0.90 & Dialogue-system tool interactions, past refs \\
708 & 0.84  & 0.588 & 0.80 & Using a tool with specific parameters \\
\bottomrule
\end{tabular}
\caption{Top A-exclusive features by Cohen's $d$ with Gemma autointerp descriptions.}
\label{tab:autointerp_top}
\end{table}

\input{sections/full_eval_table}

\section{Full Saturation Table}
\label{app:saturation_table}

Best $\Delta$ tool-correctness (max over $\alpha$) at each $|S|$ with 95\% paired-$t$
CIs, referenced from Section~\ref{sec:saturation}. ``---'' indicates no sweep coverage.

\begin{table}[H]
\centering
\resizebox{\textwidth}{!}{%
\begin{tabular}{rcccc}
\toprule
$|S|$ & \textbf{DFC A-excl} & \textbf{DFC shared} & \textbf{CC} &
\textbf{A-excl $-$ CC} \\
\midrule
1  & \textbf{+65.0} [+47.9, +82.1] & +32.5 [+7.0, +58.0]            & +47.5 [+27.0, +68.0]            & \textbf{+17.5} \\
2  & +65.0 [+47.9, +82.1]          & +17.5 [$-$1.5, +36.5]          & +57.5 [+41.5, +73.5]            & +7.5 \\
3  & +62.5 [+46.8, +78.2]          & +40.0 [+18.5, +61.5]           & +30.0 [+9.3, +50.7]             & +32.5 \\
4  & +57.5 [+39.9, +75.1]          & +37.5 [+16.2, +58.8]           & +37.5 [+20.2, +54.8]            & +20.0 \\
5  & +47.5 [+29.8, +65.2]          & +47.5 [+27.0, +68.0]           & +57.5 [+38.5, +76.5]            & $-$10.0 \\
6  & +47.5 [+29.8, +65.2]          & +22.5 [+1.4, +43.6]            & +52.5 [+32.0, +73.0]            & $-$5.0 \\
7  & +47.5 [+29.8, +65.2]          & +27.5 [+7.0, +48.0]            & ---                             & --- \\
8  & ---                            & +42.5 [+23.5, +61.5]           & +37.5 [+15.0, +60.0]            & --- \\
9  & ---                            & +45.0 [+24.6, +65.4]           & +47.5 [+25.8, +69.2]            & --- \\
10 & ---                            & +30.0 [+8.0, +52.0]            & ---                             & --- \\
33 & ---                            & ---                             & \textbf{+70.0} [+53.5, +86.5]  & --- \\
38 & ---                            & +62.5 [+43.8, +81.2]           & ---                             & --- \\
\bottomrule
\end{tabular}}
\caption{Best $\Delta$ (max over $\alpha$) at each $|S|$ with 95\% paired-$t$ CIs.
         DFC A-excl plateaus at $+47$--$65$\,pp from $|S|=1$; CC needs $|S|=33$ to
         reach $+70$\,pp. ``---'' = no sweep coverage.}
\label{tab:saturation_full}
\end{table}

\edit{%
\input{figures/steering_before_after_inline}}

\subsection{Practical Sweet Spot and $\alpha$ Tradeoff}
\label{sec:sweetspot}

\vspace{-4pt}
\enlargethispage{\baselineskip}
\begin{figure}[H]
\centering
\includegraphics[width=0.55\linewidth]{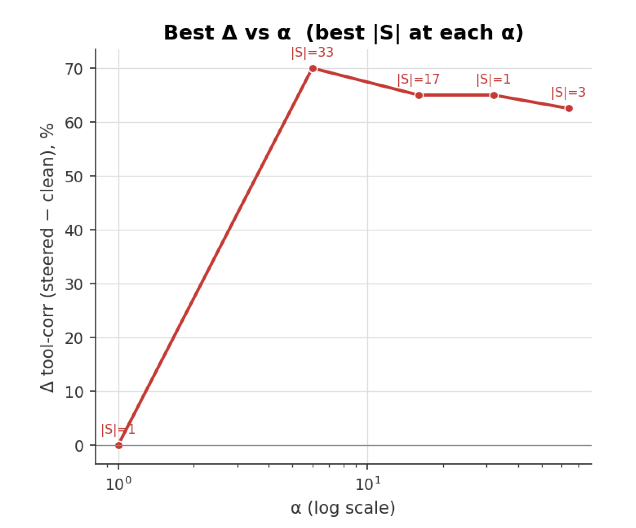}
\caption{Best $\Delta$ vs $\alpha$, pooled across DFC A-excl, shared, and CC.
         Moderate $\alpha$ ($\approx6$--$32$) is consistently near-optimal.
         Over-steering ($\alpha=64$, large $|S|$) degrades performance.}
\label{fig:tradeoff_alpha}
\end{figure}

Key operating points: DFC A-excl at $|S|=1$, $\alpha=32$ yields $+65.0$\,pp; CC at
$|S|=33$, $\alpha=6$ yields $+70.0$\,pp --- $33\times$ the neuron budget for a
$5$\,pp gain. Over-steering ($\alpha=64$, $|S|\geq9$) degrades performance (e.g.\
$|S|=33$, $\alpha=64\to+0.0$\,pp).

\subsection{Per-Condition Heatmaps}
\label{sec:heatmaps}

\vspace{-4pt}
\enlargethispage{\baselineskip}
\begin{figure}[H]
\centering
\begin{subfigure}[b]{0.48\linewidth}
  \includegraphics[width=\linewidth]{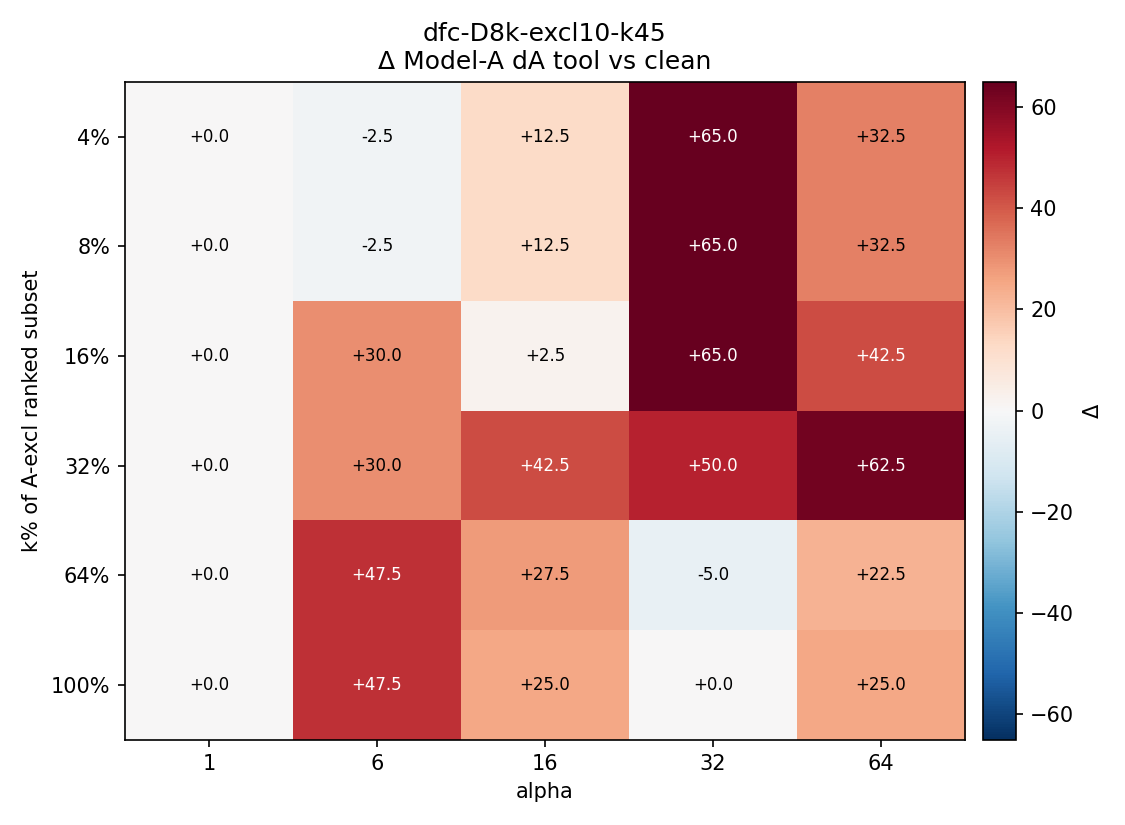}
  \caption{\texttt{dfc-D8k-excl10-k45}: sweet spot at low $|S|$, high $\alpha$.}
\end{subfigure}
\hfill
\begin{subfigure}[b]{0.48\linewidth}
  \includegraphics[width=\linewidth]{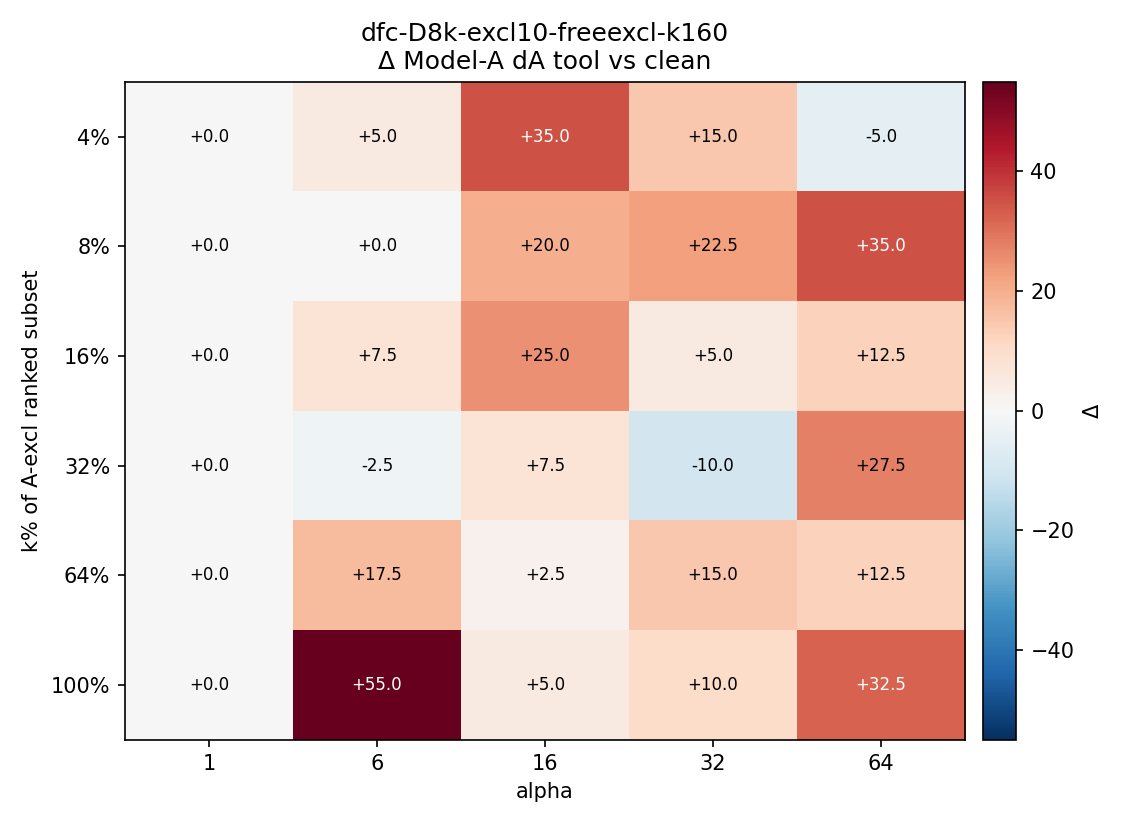}
  \caption{\texttt{dfc-D8k-excl10-freeexcl-k160}: sweet spot at high $|S|$,
           moderate $\alpha$.}
\end{subfigure}
\caption{Per-cell $\Delta$ tool-correctness heatmaps ($|S|\times\alpha$) for the two
         representative DFCs. Sparse $k=45$ peaks at targeted low-$|S|$ steering;
         dense $k=160$ benefits from broader coverage.}
\label{fig:heatmaps}
\end{figure}



\section{Models and Repository}

Our code can be found at \url{https://github.com/Antebe/model_diffing_crosscoders}. The full sweep is now available at \url{https://huggingface.co/collections/antebe1/qwen-toolrl-crosscoder}.

%% file: sections/full_eval_table.tex
\section{Full Sweep Evaluation Table}
\label{app:full_eval}

Table~\ref{tab:full_sweep_eval} reports the key hyperparameters and post-reconstruction
evaluation outcomes for all {$48$} evaluated sweep variants. We include both Model A and
Model B post-reconstruction metrics so that individual model configurations can be
cited directly in the appendix without referring back to the JSONL artifact.

\begingroup
\scriptsize
\setlength{\tabcolsep}{3pt}
\begin{longtable}{@{}p{4.8cm}lrrrrccrrrrrrrr@{}}
\caption{Full post-reconstruction evaluation table across all {$48$} sweep variants. $p$
denotes the exclusive share per model, $\lambda_{\mathrm{excl}}$ the exclusive-partition
L1 penalty, and $\lambda_{\mathrm{sh}}$ the shared-feature L1 penalty. All rows use
100 held-out ToolRL prompts.}
\label{tab:full_sweep_eval}\\
\toprule
\textbf{Model} & \textbf{Arch} & \textbf{$D$} & \textbf{$k$} & \textbf{$p$} &
\textbf{$\lambda_{\mathrm{excl}}$} & \textbf{$\lambda_{\mathrm{sh}}$} &
\textbf{A ov} & \textbf{A fmt} & \textbf{A tool} & \textbf{$\Delta$A tool} &
\textbf{B ov} & \textbf{B fmt} & \textbf{B tool} & \textbf{$\Delta$B tool} \\
\midrule
\endfirsthead

\multicolumn{15}{c}{\tablename~\thetable\ (continued)}\\
\toprule
\textbf{Model} & \textbf{Arch} & \textbf{$D$} & \textbf{$k$} & \textbf{$p$} &
\textbf{$\lambda_{\mathrm{excl}}$} & \textbf{$\lambda_{\mathrm{sh}}$} &
\textbf{A ov} & \textbf{A fmt} & \textbf{A tool} & \textbf{$\Delta$A tool} &
\textbf{B ov} & \textbf{B fmt} & \textbf{B tool} & \textbf{$\Delta$B tool} \\
\midrule
\endhead

\midrule
\multicolumn{15}{r}{Continued on next page}\\
\endfoot

\bottomrule
\endlastfoot
\texttt{cc-D16k-k160} & CrossCoder & 16384 & 160 & 0\% & 0 & $10^{-3}$ & 1 & 66 & 67 & 48 & -0.92 & 0 & 4 & 4 \\
\texttt{cc-D16k-k45} & CrossCoder & 16384 & 45 & 0\% & 0 & $10^{-3}$ & 0.55 & 51 & 52 & 33 & -0.76 & 0 & 12 & 12 \\
\texttt{cc-D16k-k90} & CrossCoder & 16384 & 90 & 0\% & 0 & $10^{-3}$ & 0.37 & 45 & 46 & 27 & -0.94 & 0 & 3 & 3 \\
\texttt{cc-D16k-nol1-k160} & CrossCoder & 16384 & 160 & 0\% & 0 & 0 & 0.26 & 42 & 42 & 23 & -0.70 & 0 & 15 & 15 \\
\texttt{cc-D16k-nol1-k45} & CrossCoder & 16384 & 45 & 0\% & 0 & 0 & 0.59 & 53 & 53 & 34 & -0.72 & 0 & 14 & 14 \\
\texttt{cc-D16k-nol1-k90} & CrossCoder & 16384 & 90 & 0\% & 0 & 0 & 0.05 & 35 & 35 & 16 & -0.70 & 0 & 15 & 15 \\
\texttt{cc-D8k-k160} & CrossCoder & 8192 & 160 & 0\% & 0 & $10^{-3}$ & 0.47 & 49 & 49 & 30 & -0.92 & 0 & 4 & 4 \\
\texttt{cc-D8k-k45} & CrossCoder & 8192 & 45 & 0\% & 0 & $10^{-3}$ & 0.76 & 58 & 59 & 40 & -0.76 & 0 & 12 & 12 \\
\texttt{cc-D8k-k90} & CrossCoder & 8192 & 90 & 0\% & 0 & $10^{-3}$ & 0.50 & 50 & 50 & 31 & -0.94 & 0 & 3 & 3 \\
\texttt{cc-D8k-nol1-k160} & CrossCoder & 8192 & 160 & 0\% & 0 & 0 & 0.21 & 41 & 40 & 21 & -0.84 & 0 & 8 & 8 \\
\texttt{cc-D8k-nol1-k45} & CrossCoder & 8192 & 45 & 0\% & 0 & 0 & 1.10 & 70 & 70 & 51 & -0.66 & 0 & 17 & 17 \\
\texttt{cc-D8k-nol1-k90} & CrossCoder & 8192 & 90 & 0\% & 0 & 0 & 0.80 & 60 & 60 & 41 & -1 & 0 & 0 & 0 \\
\texttt{dfc-D16k-excl10-freeexcl-k160} & DFC & 16384 & 160 & 10\% & 0 & $10^{-3}$ & 0.75 & 59 & 58 & 39 & -0.92 & 0 & 4 & 4 \\
\texttt{dfc-D16k-excl10-freeexcl-k45} & DFC & 16384 & 45 & 10\% & 0 & $10^{-3}$ & -0.01 & 33 & 33 & 14 & -0.98 & 0 & 1 & 1 \\
\texttt{dfc-D16k-excl10-freeexcl-k90} & DFC & 16384 & 90 & 10\% & 0 & $10^{-3}$ & 0.69 & 57 & 56 & 37 & -0.62 & 0 & 19 & 19 \\
\texttt{dfc-D16k-excl10-k160} & DFC & 16384 & 160 & 10\% & $10^{-3}$ & $10^{-3}$ & 0.60 & 52 & 54 & 35 & -0.90 & 0 & 5 & 5 \\
\texttt{dfc-D16k-excl10-k45} & DFC & 16384 & 45 & 10\% & $10^{-3}$ & $10^{-3}$ & 0.54 & 50 & 52 & 33 & -0.86 & 0 & 7 & 7 \\
{\texttt{dfc-D16k-excl10-k90}} & {DFC} & {16384} & {90} & {10\%} & {$10^{-3}$} & {$10^{-3}$} & {0.54} & {50} & {52} & {27} & {-0.70} & {0} & {15} & {15} \\
\texttt{dfc-D16k-excl3-freeexcl-k160} & DFC & 16384 & 160 & 3\% & 0 & $10^{-3}$ & 0.49 & 49 & 50 & 31 & -0.94 & 0 & 3 & 3 \\
\texttt{dfc-D16k-excl3-freeexcl-k45} & DFC & 16384 & 45 & 3\% & 0 & $10^{-3}$ & 0.49 & 49 & 50 & 31 & -0.88 & 0 & 6 & 6 \\
\texttt{dfc-D16k-excl3-freeexcl-k90} & DFC & 16384 & 90 & 3\% & 0 & $10^{-3}$ & 0.23 & 41 & 41 & 22 & -0.82 & 0 & 9 & 9 \\
\texttt{dfc-D16k-excl3-k160} & DFC & 16384 & 160 & 3\% & $10^{-3}$ & $10^{-3}$ & 0.56 & 52 & 52 & 33 & -0.84 & 0 & 8 & 8 \\
\texttt{dfc-D16k-excl3-k45} & DFC & 16384 & 45 & 3\% & $10^{-3}$ & $10^{-3}$ & 0.28 & 42 & 43 & 24 & -0.88 & 0 & 6 & 6 \\
\texttt{dfc-D16k-excl3-k90} & DFC & 16384 & 90 & 3\% & $10^{-3}$ & $10^{-3}$ & 0.26 & 42 & 42 & 23 & -0.86 & 0 & 7 & 7 \\
\texttt{dfc-D16k-excl5-freeexcl-k160} & DFC & 16384 & 160 & 5\% & 0 & $10^{-3}$ & 0.74 & 58 & 58 & 39 & -0.94 & 0 & 3 & 3 \\
\texttt{dfc-D16k-excl5-freeexcl-k45} & DFC & 16384 & 45 & 5\% & 0 & $10^{-3}$ & 0.44 & 48 & 48 & 29 & -0.88 & 0 & 6 & 6 \\
\texttt{dfc-D16k-excl5-freeexcl-k90} & DFC & 16384 & 90 & 5\% & 0 & $10^{-3}$ & 0.33 & 43 & 45 & 26 & -0.86 & 0 & 7 & 7 \\
\texttt{dfc-D16k-excl5-k160} & DFC & 16384 & 160 & 5\% & $10^{-3}$ & $10^{-3}$ & 0.49 & 49 & 50 & 31 & -0.92 & 0 & 4 & 4 \\
\texttt{dfc-D16k-excl5-k45} & DFC & 16384 & 45 & 5\% & $10^{-3}$ & $10^{-3}$ & 0.37 & 45 & 46 & 27 & -0.94 & 0 & 3 & 3 \\
\texttt{dfc-D16k-excl5-k90} & DFC & 16384 & 90 & 5\% & $10^{-3}$ & $10^{-3}$ & 0.49 & 49 & 50 & 31 & -0.94 & 0 & 3 & 3 \\
\texttt{dfc-D8k-excl10-freeexcl-k160} & DFC & 8192 & 160 & 10\% & 0 & $10^{-3}$ & 1.49 & 83 & 83 & 64 & -1 & 0 & 0 & 0 \\
\texttt{dfc-D8k-excl10-freeexcl-k45} & DFC & 8192 & 45 & 10\% & 0 & $10^{-3}$ & 0.27 & 41 & 43 & 24 & -0.88 & 0 & 6 & 6 \\
\texttt{dfc-D8k-excl10-freeexcl-k90} & DFC & 8192 & 90 & 10\% & 0 & $10^{-3}$ & 0.56 & 52 & 52 & 33 & -0.82 & 0 & 9 & 9 \\
\texttt{dfc-D8k-excl10-k160} & DFC & 8192 & 160 & 10\% & $10^{-3}$ & $10^{-3}$ & 0.32 & 44 & 44 & 25 & -0.84 & 0 & 8 & 8 \\
\texttt{dfc-D8k-excl10-k45} & DFC & 8192 & 45 & 10\% & $10^{-3}$ & $10^{-3}$ & 0.70 & 56 & 57 & 38 & -0.60 & 0 & 20 & 20 \\
\texttt{dfc-D8k-excl10-k90} & DFC & 8192 & 90 & 10\% & $10^{-3}$ & $10^{-3}$ & 0.60 & 54 & 53 & 34 & -0.96 & 0 & 2 & 2 \\
\texttt{dfc-D8k-excl3-freeexcl-k160} & DFC & 8192 & 160 & 3\% & 0 & $10^{-3}$ & 0.68 & 56 & 56 & 37 & -0.88 & 0 & 6 & 6 \\
\texttt{dfc-D8k-excl3-freeexcl-k45} & DFC & 8192 & 45 & 3\% & 0 & $10^{-3}$ & 0.47 & 49 & 49 & 30 & -0.88 & 0 & 6 & 6 \\
\texttt{dfc-D8k-excl3-freeexcl-k90} & DFC & 8192 & 90 & 3\% & 0 & $10^{-3}$ & 0.10 & 36 & 37 & 18 & -0.82 & 0 & 9 & 9 \\
\texttt{dfc-D8k-excl3-k160} & DFC & 8192 & 160 & 3\% & $10^{-3}$ & $10^{-3}$ & 0.50 & 50 & 50 & 31 & -0.90 & 0 & 5 & 5 \\
\texttt{dfc-D8k-excl3-k45} & DFC & 8192 & 45 & 3\% & $10^{-3}$ & $10^{-3}$ & 0.32 & 44 & 44 & 25 & -0.98 & 0 & 1 & 1 \\
\texttt{dfc-D8k-excl3-k90} & DFC & 8192 & 90 & 3\% & $10^{-3}$ & $10^{-3}$ & 0.32 & 44 & 44 & 25 & -0.88 & 0 & 6 & 6 \\
\texttt{dfc-D8k-excl5-freeexcl-k160} & DFC & 8192 & 160 & 5\% & 0 & $10^{-3}$ & 0.53 & 51 & 51 & 32 & -0.92 & 0 & 4 & 4 \\
\texttt{dfc-D8k-excl5-freeexcl-k45} & DFC & 8192 & 45 & 5\% & 0 & $10^{-3}$ & 0.71 & 57 & 57 & 38 & -0.96 & 0 & 2 & 2 \\
\texttt{dfc-D8k-excl5-freeexcl-k90} & DFC & 8192 & 90 & 5\% & 0 & $10^{-3}$ & 0.92 & 64 & 64 & 45 & -0.94 & 0 & 3 & 3 \\
\texttt{dfc-D8k-excl5-k160} & DFC & 8192 & 160 & 5\% & $10^{-3}$ & $10^{-3}$ & -0.22 & 26 & 26 & 7 & -0.96 & 0 & 2 & 2 \\
\texttt{dfc-D8k-excl5-k45} & DFC & 8192 & 45 & 5\% & $10^{-3}$ & $10^{-3}$ & 0.35 & 45 & 45 & 26 & -0.94 & 0 & 3 & 3 \\
\texttt{dfc-D8k-excl5-k90} & DFC & 8192 & 90 & 5\% & $10^{-3}$ & $10^{-3}$ & 0.55 & 51 & 52 & 33 & -0.74 & 0 & 13 & 13 \\
\end{longtable}
\endgroup

%% file: figures/steering_before_after_inline.tex
%

\usetikzlibrary{positioning}

\definecolor{promptblue}{HTML}{355C8C}
\definecolor{panelpurple}{HTML}{7F7DB6}
\definecolor{beforebadge}{HTML}{A11D1D}
\definecolor{afterbadge}{HTML}{2F6F20}
\definecolor{outputpurple}{HTML}{6D47CC}
\definecolor{softbg}{HTML}{F4F4FB}
\definecolor{shadowgray}{HTML}{B8BEC7}
\definecolor{toolorange}{HTML}{D97706}
\definecolor{thinkblue}{HTML}{1D4ED8}
\definecolor{callgreen}{HTML}{4D9F0C}

\tikzset{
  sba_panel/.style={draw=panelpurple!70!black, rounded corners=6pt, line width=0.8pt, fill=softbg},
  sba_box/.style={draw=black!60, rounded corners=3pt, line width=0.6pt, fill=white, inner sep=7pt},
  sba_header/.style={fill=panelpurple, rounded corners=6pt, text=white,
    font=\bfseries\sffamily\scriptsize, align=center,
    inner xsep=10pt, inner ysep=5pt, anchor=north west},
  sba_badge/.style={rounded corners=3pt, text=white,
    font=\bfseries\sffamily\footnotesize,
    inner xsep=8pt, inner ysep=2pt, anchor=north west},
}

\begin{figure}[tbp]
\centering
\resizebox{\textwidth}{!}{%
\begin{tikzpicture}[x=1cm,y=1cm]

\fill[shadowgray!55, rounded corners=2pt] (0.14,-0.18) rectangle (15.84,-1.62);
\draw[rounded corners=2pt, line width=0.7pt, draw=black!65, fill=white]
   (0,-0.05) rectangle (15.7,-1.5);
\node[fill=promptblue, rounded corners=2pt, text=white,
      font=\bfseries\sffamily\small, inner xsep=8pt, inner ysep=4pt,
      anchor=north west] at (0.10,-0.10) {PROMPT};
\node[anchor=west, text width=13.6cm, align=left, font=\ttfamily\footnotesize]
   at (2,-0.38)
   {ToolRL hold-out (\texttt{prompt\_index=2}): system prompt listing seven list/string utilities and asking the model to emit \texttt{<think>}/\texttt{<tool\_call>}/\texttt{<response>}.};
\node[anchor=west, text width=13.7cm, align=left,
      font=\itshape\sffamily\footnotesize, text=toolorange] at (1.85,-1.1)
   {Available tools: \texttt{flatten\_list}, \texttt{split\_list}, \texttt{get\_range}, \texttt{reverse\_words}, \texttt{reverse\_string}, \texttt{merge\_dictionaries}, \texttt{is\_subset}.};

\fill[shadowgray!55, rounded corners=6pt] (0.22,-2.18) rectangle (7.62,-9.78);
\fill[shadowgray!55, rounded corners=6pt] (8.22,-2.18) rectangle (15.62,-9.78);
\draw[sba_panel] (0,-2.0)  rectangle (7.4,-10.15);
\draw[sba_panel] (8,-2.0)  rectangle (15.4,-10.15);

\node[sba_header] at (0.18,-2.05)
   {TOOLRL-QWEN2.5-3B \\ + DFC RECONSTRUCTION (CLEAN)};
\node[sba_badge, fill=beforebadge] at (0.35,-3.05) {BEFORE STEERING};
\node[sba_badge, fill=beforebadge] at (0.35,-3.85)
   {NO \texttt{<tool\_call>} EMITTED};
\node[sba_box, anchor=north west, text width=6.18cm, align=left,
      font=\ttfamily\scriptsize] at (0.35,-4.62)
  {%
  \textcolor{thinkblue}{\texttt{<think>}} The user's request is not\\
  clear from the provided information. I\\
  need more details about what task or\\
  information they are looking for.\\
  \textcolor{thinkblue}{\texttt{</think>}}\\[3pt]
  \texttt{<response>} Could you please\\
  provide more details about the task\\
  you want to accomplish? Are you\\
  looking for information, performing a\\
  calculation, or need help with a\\
  specific process? \texttt{</response>}\\[5pt]
  $\Rightarrow$ \texttt{format\_accuracy}=False\\
  $\Rightarrow$ \texttt{tool\_correctness}=False};

\node[sba_header] at (8.18,-2.05)
   {SAME MODEL \\ + STEER FEAT \#136, $|S|{=}1$, $\alpha{=}32$};
\node[sba_badge, fill=afterbadge] at (8.35,-3.05) {AFTER STEERING};
\node[sba_badge, fill=outputpurple] at (8.35,-3.85)
   {STRUCTURED TOOL CALL};
\node[sba_box, anchor=north west, text width=6.20cm, align=left,
      font=\ttfamily\scriptsize] at (8.35,-4.62)
  {%
  \textbf{User:} I have a list of numbers\\
  {[1,2,3,4,5]}. How can I split it into\\
  chunks of 2?\\[3pt]
  \textcolor{thinkblue}{\texttt{<think>}} The user wants to split\\
  the given list into chunks of 2. We can\\
  use the \texttt{'split\_list'} function to\\
  achieve this. The input list is\\
  {[1,2,3,4,5]} and the chunk size is 2.\\
  \textcolor{thinkblue}{\texttt{</think>}}\\[3pt]
  \textcolor{callgreen}{\texttt{<tool\_call>}}\\
  \texttt{\{"name":"split\_list",}\\
  \texttt{\ "parameters":\{"lst":[1,2,3,4,5],}\\
  \texttt{\ \ "chunk\_size":2\}\}}\\
  \textcolor{callgreen}{\texttt{</tool\_call>}}\\[3pt]
  $\Rightarrow$ \texttt{format\_accuracy}=True\\
  $\Rightarrow$ \texttt{tool\_correctness}=True};

\end{tikzpicture}%
}%
\caption{{\textbf{Tool-use behaviour before vs.\ after a single A-exclusive steering vector.}
Both panels show the same ToolRL-Qwen2.5-3B passed through the same DFC
(\texttt{dfc-D8k-excl10-k45}, layer~13). \textbf{Left:} standard reconstruction --- the
model asks for clarification and never emits a \texttt{<tool\_call>}
($\texttt{format\_accuracy}=$False, $\texttt{tool\_correctness}=$False).
\textbf{Right:} the same model with an additive delta on the single
highest-Cohen's-$d$ A-exclusive feature (index~$136$), $|S|=1$, $\alpha=32$ --- the
model now infers a plausible user goal, reasons through it inside \texttt{<think>}, and
emits a syntactically valid \texttt{<tool\_call>} naming \texttt{split\_list} with the
correct argument schema ($\texttt{format\_accuracy}=$True, $\texttt{tool\_correctness}=$True).}}
\label{fig:toolrl_before_after}
\end{figure}